\documentclass{article} % For LaTeX2e
\usepackage{iclr2016_conference,times}
\usepackage{hyperref}
\usepackage{url}
\usepackage{amsmath,amsfonts,amsthm, amssymb}
\usepackage{graphicx}
\usepackage{subfig}
\usepackage{multirow}
\usepackage{float}
\title{Convolutional Neural Networks With Low-rank Regularization}

\author{Cheng Tai$^1$, Tong Xiao$^2$, Yi Zhang$^3$, Xiaogang Wang$^2$, Weinan E$^1$ \\
$^1$The Program in Applied and Computational Mathematics, Princeton University \\
$^2$Department of Electronic Engineering, The Chinese University of Hong Kong \\
$^3$Department of Electrical Engineering and Computer Science, University of Michigan, Ann Arbor \\
\texttt{\{chengt,weinan\}@math.princeton.edu}; \hspace{0.15cm}
\texttt{yeezhang@umich.edu} \\
\texttt{\{xiaotong,xgwang\}@ee.cuhk.edu.hk}
}

\newcommand{\R}{\mathbb{R}}

\renewcommand{\H}{\mathcal{H}}
\newcommand{\V}{\mathcal{V}}
\newcommand{\W}{\mathcal{W}}
\newcommand{\Z}{\mathcal{Z}}

\newcommand{\specialcell}[2][c]{\begin{tabular}[#1]{@{}c@{}}#2\end{tabular}}
\iclrfinalcopy % Uncomment for camera-ready version

\newtheorem{theorem}{Theorem}

\begin{document}

\maketitle
\begin{abstract}
Large CNNs have delivered impressive performance in various computer vision applications. But the storage and computation requirements make it problematic for deploying these models on mobile devices. Recently, tensor decompositions have been used for speeding up CNNs. In this paper, we further develop the tensor decomposition technique. We propose a new algorithm for computing the low-rank tensor decomposition for removing the redundancy in the convolution kernels. The algorithm finds the exact global optimizer of the decomposition and is more effective than iterative methods. Based on the decomposition, we further propose a new method for training low-rank constrained CNNs from scratch. Interestingly, while achieving a significant speedup, sometimes the low-rank constrained CNNs delivers significantly better performance than their non-constrained counterparts. On the  CIFAR-10 dataset, the proposed low-rank NIN model achieves $91.31\%$ accuracy (without data augmentation), which also improves upon state-of-the-art result.  We evaluated the proposed method on CIFAR-10 and ILSVRC12 datasets for a variety of modern CNNs, including AlexNet, NIN, VGG and GoogleNet with success. For example, the forward time of VGG-16 is reduced by half while the performance is still comparable. Empirical success suggests that low-rank tensor decompositions can be a very useful tool for speeding up large CNNs.

\end{abstract}

\section{Introduction}
Over the course of three years, CNNs have revolutionized computer vision, setting new performance standards in many important applications, see e.g., \citet{krizhevsky2012imagenet,farabet2013learning,long2014fully}. The breakthrough has been made possible by the abundance of training data, the deployment of new computational hardware (most notably, GPUs and CPU clusters) and large models. These models typically require a huge number of parameters ($10^7\sim 10^9$) to achieve state-of-the-art performance, and may take weeks to train even with high-end GPUs.  On the other hand, there is a growing interest in deploying CNNs to low-end mobile devices. On such processors, the computational cost of applying the model becomes problematic, let alone training one, especially when real-time operation is needed. Storage of millions of parameters also complicates the deployment. Modern CNNs would  find many more applications if both the computational cost and the storage requirement could be significantly reduced.

There are only a few recent works for speeding up CNNs. \citet{denton2014exploiting} proposed some low-rank approximation and clustering schemes for the convolutional kernels. They achieved 2x speedup for a single convolutional layer with 1\% drop in classification accuracy. \cite{jaderberg2014speeding} suggested using different tensor decomposition schemes, reporting a 4.5x speedup with 1\% drop in accuracy in a text recognition application. \citet{lebedev2014speeding} further explored the use of CP decomposition to approximate the convolutional kernels. \citet{vanhoucke2011improving} showed that using 8-bit quantization of the parameters can result in significant speedup with minimal loss of accuracy. This method can be used in conjunction with low-rank approximations to achieve further speedup.

As convolution operations constitute the bulk of all computations in CNNs, simplifying the convolution layer would have a direct impact on the overall speedup. The convolution kernels in a typical CNN is a 4D tensor. The key observation is that there might be a significant amount of redundancy in the tensor. Ideas based on tensor decomposition seem to be a particularly promising way to remove the redundancy as suggested by some previous works.

In this paper, we further develop the tensor decomposition idea. Our method is based on \citet{jaderberg2014speeding}, but has several significant improvements. The contributions are summarized as follows:
\begin{itemize}
\item A new algorithm for computing the low-rank tensor decomposition.  Low-rank tensor decompositions are non-convex problems and difficult to compute in general, \citet{jaderberg2014speeding} use iterative schemes to get an approximate local solution. But we find that the particular form of low-rank decomposition in \citep{jaderberg2014speeding} has an exact closed form solution which is the  global optimum. Hence we obtain the best data-independent approximation. Furthermore, computing the exact solution is much more effective than iterative schemes. As the tensor decomposition is the most important step in approximating CNNs, being able to obtain an exact solution efficiently thus provides great advantages.
\item A new method for training low-rank constrained CNNs from scratch. Most previous works only focus on improving testing time computation cost. This is achieved by approximating and fine-tuning a pre-trained network. Based on the low-rank tensor decomposition, we find that the convolutional kernels can be parameterized in a way that naturally enforces the low-rank constraint. Networks parameterized in this low-rank constrained manner have more layers than their non-constrained counterparts. While it is widely observed that deeper networks are harder to train, we are able to train very deep low-rank constrained CNNs with more than 30 layers with the help of a recent training technique called \textit{batch normalization} \citet{ioffe2015batch}.
\item Evaluation on large networks.  Previous experiments in \citet{jaderberg2014speeding} and \citet{denton2014exploiting} give some promises of the effectiveness of low-rank approximations. But these methods have not been tested extensively for large models and generic datasets. Moreover, as iterative methods are used to find the approximation, bad local minima may hurt performance. In this paper, we test the proposed method for various state-of-the-art CNN models, including NIN \citep{2013arXiv1312.4400L}, AlexNet \citep{krizhevsky2012imagenet}, VGG \citep{simonyan2014very} and GoogleNet \citep{szegedy2014going}. The datasets used include CIFAR-10 and ILSVRC12. We achieved significant speedups for these models with comparable or even better performance. Success on a variety of CNN models give strong evidence that  low-rank tensor decomposition can be a very useful tool for simplifying and improving deep CNNs.
\end{itemize}

Our numerical experiments show that significant speedup can be achieved with minimal loss of performance, which is consistent with previously reported results. Surprisingly, while all previous efforts report a slight decrease or no change in performance, we found a significant increase of classification accuracy in some cases. In particular, on the CIFAR-10 dataset, we achieve 91.31\% classification accuracy (without data augmentation) with the low-rank NIN model, which improves upon not only the original NIN but also upon state-of-the-art results on this dataset. We are not aware of significant improvements with low-rank approximations being reported in the previous literature.

The rest of the paper is organized as follows. We discuss some related work in section 2. We then introduce our decomposition scheme in section 3. Results with typical networks including AlexNet, NIN, VGG  and GoogleNet on CIFAR10 and ILSVRC12 datasets are reported in section 4. We conclude with the summary and discussion in Section 5.

\section{Related Work}
Using low-rank filters to accelerate convolution has a long history. Classic examples include high dimensional DCT and wavelet systems constructed from 1D wavelets using tensor products. In the context of dictionary learning, learning separable 1D filters was suggested by \citet{rigamonti2013learning}.

More specific to CNNs, there are two works that are most related to ours: \citet{jaderberg2014speeding,lebedev2014speeding}. For \citet{jaderberg2014speeding}, in addition to the improvements summarized in the previous section, there is another difference in the approximation stage. In \citet{jaderberg2014speeding}, the network is approximated layer by layer. After one layer is approximated by the low-rank filters, the parameters of that layer are fixed, and the layers above are fine-tuned based on a reconstruction error criterion. Our scheme fine-tunes the entire network simultaneously using a discriminative criterion. While \citet{jaderberg2014speeding} reported that discriminative fine-tuning was inefficient for their scheme, we found that it works very well in our case.

In \citet{lebedev2014speeding}, CP decomposition of the kernel tensors is proposed. \citet{lebedev2014speeding} used non-linear least squares to compute the CP decomposition. It is also based on the tensor decomposition idea, but our decomposition is based on a different scheme and has some numerical advantages.  For the CP decomposition, finding the best low-rank approximation is an ill-posed problem, and the best rank-$K$ approximation may not exist in the general case, regardless the choice of norm \citep{deSilva:2008bg}. But for the proposed scheme, the decomposition always exists, and we have an exact closed form solution for the decomposition. In principle, both the CP decomposition scheme and the proposed scheme can be used to train CNNs from scratch. In the CP decomposition, one convolutional layer is replaced with four convolutional layers. Although the effective depth of the network remains the same, it makes optimization much harder as the gradients of the inserted layers are prone to explosion. Because of this, application of this scheme to larger and deeper models is still problematic due to numerical issues.

Lastly, different from both, we consider more and much larger models, which is more challenging. Thus our results provide strong evidence that low-rank approximations can be applicable to a variety of state-of-the-art models.

\section{Method}
In line with the method in \citet{jaderberg2014speeding}, the proposed tensor decomposition scheme is based on a conceptually simple idea: replace the 4D convolutional kernel with two consecutive kernels with a lower rank. In the following, we introduce the details of the decomposition and the algorithms of using the decomposition to approximate a pre-trained network and to train a new one.

\subsection{Approximation of a pre-trained CNN}
Formally, a convolutional kernel in a CNN is a 4D tensor $\W\in \R^{N\times d\times d\times C}$, where $N, C$ are the numbers of the output and input feature maps respectively and $d$ is the spatial kernel size. We also view $\W$ as an 3D filter array and use notation $\W_n\in \R^{d\times d\times C}$ to represent the $n$-th filter. Let $\Z\in\R^{X\times Y\times C}$ be the input feature map. The output feature map $\mathcal{F}=(\mathcal{F}_1,\cdots,\mathcal{F}_N)$ is defined as
\[
	\mathcal{F}_n(x,y) = \sum_{i=1}^C\sum_{x'=1}^{X}\sum_{y'=1}^{Y} \Z^c(x',y')\W_n^c(x-x',y-y'),
\]
where the superscript is the index of the channels.

The goal is to find an approximation $\tilde{\W}$ of $\W$ that facilitates more efficient computation while maintaining the classification accuracy of the CNN. We propose the following scheme:
\begin{equation}
  \tilde{\W}_n^c = \sum_{k=1}^K \H_n^k(\V_k^c)^T,
\label{eq:scheme}
\end{equation}
where $K$ is a hyper-parameter controlling the rank, $\H\in \R^{N\times 1\times d\times K}$ is the horizontal filter, $\V \in \R^{K\times d\times 1\times C}$ is the vertical filter (we have slightly abused the notations to make them concise, $\H_n^k$ and $\V_k^c$ are both vectors in $\R^d$). Both $\H$ and $\V$ are learnable parameters.

With this form, the convolution becomes:
\begin{equation}
 \tilde{\W}_n*\Z = \sum_{c=1}^C \sum_{k=1}^K \H_n^k(\V_k^c)^T*\Z^c
 = \sum_{k=1}^K \H_n^k*\left( \sum_{c=1}^C \V_k^c * \Z^c\right)
\end{equation}
 The intuition behind this approximation scheme is to exploit the redundancy that exist both in the spatial dimensions and across channels. Note the convolutions in the above equation are all one dimensional in space.

 We can estimate the reduction in computation with this scheme. Direct convolution by definition requires $\mathcal{O}(d^2NCXY)$ operations. In the above scheme, the computational cost associated with the vertical filters is $\mathcal{O}(dKCXY)$ and with horizontal filters $\mathcal{O}(dNKXY)$, giving a total computational cost of $\mathcal{O}(dK(N+C)XY)$. Acceleration can be achieved if we choose $K < \frac{dNC}{N+C}$. In principle, if $C \ll N$, which is typical in the first layer of a CNN, the acceleration is about $d$ times.

We learn the approximating parameters $H$ and $V$ by a two-step strategy. In the first step, we approximate the convolution kernel $\W$ in each layer by minimizing $\|\tilde{\W}-\W\|_F$ (index of the layers are omitted for notation simplicity). Note that this step can be done in parallel as there is no inter-layer dependence. Then we fine-tune the whole CNN based on the discriminative criterion of restoring classification accuracy.
\begin{figure}
\centering
\subfloat[]{\includegraphics[width=4.5cm]{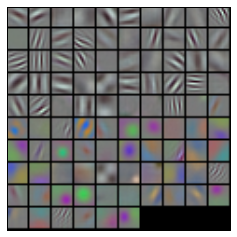}}\;
\subfloat[]{\includegraphics[height=4.5cm]{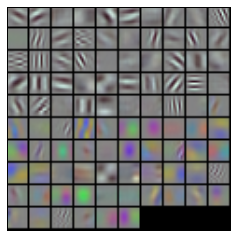}}\;
\subfloat[]{\includegraphics[height=4.5cm]{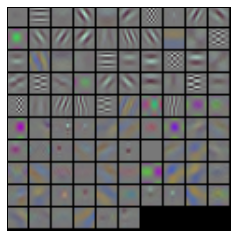}}
\caption{(a) Filters in the first layer in AlexNet. (b) Low-rank approximation using the proposed schemes with $K=8$, corresponding to $3.67\times$ speedup for this layer. Note the low-rank approximation captures most of the information, including the directionality of the original filters. (c) Low-rank filters trained from scratch with $K=8$.}
\label{fig:alex}
\end{figure}

\subsection{Algorithm}
Based on the approximation criterion introduced in the previous section, the objective function to be minimized is:
\begin{equation}
  (P1)   \quad\quad    E_1(\H,\V): = \sum_{n,c} \|\W_n^c - \sum_{k=1}^K \H_n^k (\V_k^c)^T\|_F^2.
\end{equation}
This minimization problem has a closed form solution. This is summarized in the following theorem and the proof can be found in the appendix. The theorem  gives us an efficient algorithm for computing the exact decomposition.
\begin{theorem}
Define the following bijection that maps a tensor to a matrix $\mathcal{T}:\R^{C\times d\times d\times N}  \mapsto \R^{Cd\times dN}
$, tensor element $(i_1,i_2,i_3,i_4)$ maps to $(j_1,j_2)$, where
\[
j_1 = (i_1-1)d+i_2, \quad j_2=(i_4-1)d+i_3.
\]
Define $W := \mathcal{T}[\mathcal{W}]$. Let $W=UDQ^T$ be the Singular Value Decomposition (SVD) of $W$. Let
 \begin{equation}
 \begin{aligned}
      \hat{\V}_k^c(j) &= U_{(c-1)d+j,k} \sqrt{D_{k,k}} \\
      \hat{\H}_n^k(j) &= Q_{(n-1)d+j,k}\sqrt{D_{k,k}},
 \end{aligned}
 \label{eq:soln1}
 \end{equation}
then $(\hat{\H},\hat{\V})$ is a solution to $(P1)$.
\end{theorem}

Because of this Theorem, we call the filters $(\H,\V)$ low-rank constrained filters. Note that the solution to $(P1)$ is not unique. Indeed, if $(\H,\V)$ is a solution, then $(\alpha \H, 1/\alpha \V)$ is also a solution for any $\alpha \neq 0$, but these solutions are equivalent in our application. An illustration of the closed-form approximation is shown in Figure \ref{fig:alex}.

A different criterion which uses the data distribution is proposed in \citet{denton2014exploiting}. But minimization for this criterion is NP-hard. The proof is also included in the appendix.

The algorithm provided by the above theorem is extremely fast. In our experiments, it completes in less than 1 second for most modern CNNs (AlexNet, VGG, GoogLeNet), as they have small convolutional kernels. Iterative algorithms (\citet{denton2014exploiting,jaderberg2014speeding} take much longer, especially with the data-dependent criterion. In addition, iterative algorithms often lead to bad local minimum, which leads to inferior performance even after fine-tuning. The proposed algorithm solves this issue, as it directly provides the global minimum, which is the best data-independent approximation.
Numerical demonstrations are given in section 4.

\subsection{Training Low-rank Constrained CNN From Scratch}
\begin{figure}
\centering
\includegraphics[height=4.5cm]{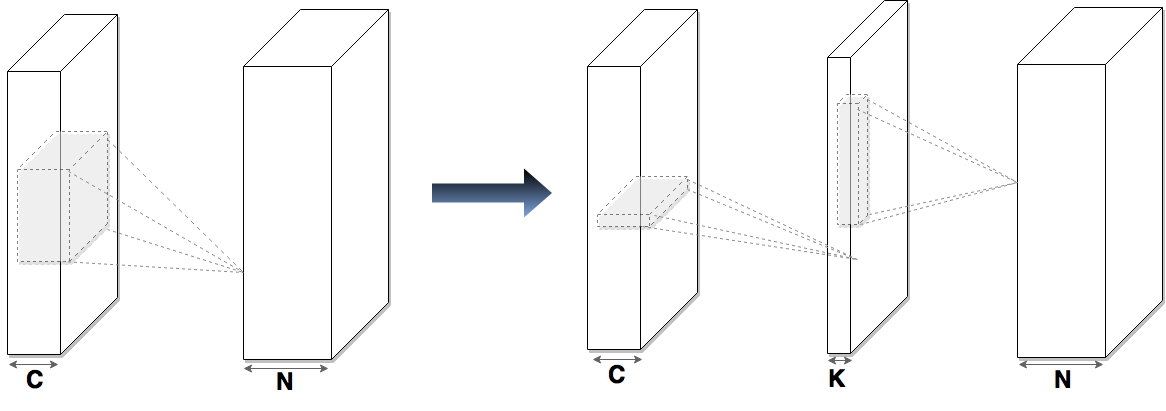}

\caption{The proposed parametrization for low-rank regularization. Left: The original convolutional layer. Right: low-rank constraint convolutional layer with rank-K.}
\label{fig:archi}
\end{figure}

Using the above scheme to train a new CNN from scratch is conceptually straightforward. Simply parametrize the convolutional to be of the form in \eqref{eq:scheme}, and the rest is not very different from training a non-constrained CNN. Here $\H$ and $\V$ are the trainable parameters. As each convolutional layer is parametrized as the composition of two convolutional layers, the resulting CNN has more layers than the original one. Although the effective depth of the new CNN is not increased, the additional layers make numerical optimization much more challenging due to exploding and vanishing gradients, especially for large networks. To handle this problem, we use a recent technique called Batch Normalization (BN) \citep{ioffe2015batch}. BN transform normalizes the activations of the internal hidden units, hence it can be an effective way to deal with the exploding or vanishing gradients. It is reported in \citet{ioffe2015batch} that deeper networks can be trained with BN successfully, and larger learning rates can be used. Empirically, we find BN effective in learning the low-rank constrained networks. An illustration of transformation of a original convolutional layer into a low-rank constraint one is in Figure \ref{fig:archi}. More details can be found in the numerical experiments section.

\section{Experiments}
In this section, we evaluate the proposed scheme on the CIFAR-10 and the ILSVRC12 datasets with several CNN models.

\subsection{CIFAR-10}
CIFAR-10 dataset is small by today's standard, but it is a good testbed for new ideas.
We deploy two models as baseline models; one is a customized CNN and the other is the NIN model. We compare their performance with their corresponding low-rank constrained versions. All models on this dataset are learned from scratch.
\begin{table}
\small
\caption{Network structure for CIFAR-10}
\label{tb:cifar_net}
\begin{minipage}[t]{.48 \linewidth}
\begin{tabular}[t]{ |c|c|c| }
\hline
Layer name & CNN &Low-rank CNN \\
\hline
conv1 &$5 \times5 \times 192$ & $K_1=12$\\
\hline
conv2 &$5\times5\times128$ & $K_2=64$\\
\hline
conv3 &$5\times5\times256$ & $K_3=128$ \\
\hline

fc1&  \multicolumn{2}{|c|}{$2304\times 512$} \\
\hline
fc2&  \multicolumn{2}{|c|}{$512\times 10$} \\
\hline
\end{tabular}
\end{minipage}\hfill
\begin{minipage}[t]{.48 \textwidth}
\begin{tabular}[t]{ |c|c|c| }
\hline
Layer name & NIN & Low-rank NIN \\
\hline
conv1 &$5\times5\times192$ & $K_1=10$\\
\hline
conv2,3 &\multicolumn{2}{|c|}{$1\times 1\times 160,1\times1\times96$} \\

\hline
conv4 & $5\times 5\times 192$ & $K_2=51$ \\
\hline
conv5,6 &\multicolumn{2}{|c|}{$1\times 1\times 192,1\times 1\times 192$} \\
\hline
conv7 & \multicolumn{2}{|c|}{$3\times 3\times 192$}\\
\hline
conv8,9 & \multicolumn{2}{|c|}{$1\times 1\times 192,1\times 1\times 10$}\\
\hline
\end{tabular}
\end{minipage}
\end{table}
\begin{table}
\small
\caption{CIFAR-10 performance}
\label{tb:cifar_performance}
\begin{center}
\begin{tabular}{llll}
\multicolumn{1}{c}{\bf METHOD}  & \multicolumn{1}{c}{\bf WITHOUT AUG.}
&\multicolumn{1}{c}{\bf WITH AUG.}
&\multicolumn{1}{c}{\bf SPEEDUP}
\\
\hline\noalign{\smallskip}
CNN (ours) & 15.12\% & 12.62\%  &1$\times$\\
Low-rank CNN (ours) & 14.50\% & 13.10\%  &2.9$\times$\\
CNN + Dropout (ours) & 13.90\% & 12.29\% & $0.96\times$\\
Low-rank CNN + Dropout (ours) & 13.81\% & 11.41\% & $2.8\times$\\
\hline\noalign{\smallskip}
NIN (ours) & 10.12\% & 8.19\% & 1$\times$\\
Low-rank NIN (ours) & {\bf 8.69\%} & {\bf 6.98\%} &1.5$\times$\\
\hline\noalign{\smallskip}
CNN + Maxout \citep{goodfellow2013maxout} & 11.68\% & 9.38\% &-\\
NIN \citep{2013arXiv1312.4400L} & 10.41\% & 8.81\% &-\\
CNN \citep{srivastava2014dropout} & 12.61\% & - &-\\
NIN + APL units \citep{agostinelli2014learning} & 9.59\% &  7.51\% &-\\
\end{tabular}
\end{center}
\end{table}

The configurations of the baseline models and their low-rank counterparts are outlined in Table \ref{tb:cifar_net}. We substitute every single convolutional layer in the baseline models with two convolutional layers with parameter $K$ introduced in the previous section. All other specifications of the network pairs are the same.  Rectified Linear Unit (ReLU) is applied to every layer except for the last one.
Our implementation of the NIN model is slightly different from the one introduced in \citet{2013arXiv1312.4400L}. We did not replace the $3 \times 3$ convolutional layer because this layer only constitutes a small fraction of the total execution time. Hence the efficiency gain of factorizing this layer is small.

The networks are trained with back propagation to optimize the multinomial logistic regression objective. The batch size is $100$. The learning learning rate is initially set to $0.01$ and decreases by a factor of $10$ every time the validation error stops decreasing. Some models have dropout units with probability $0.25$ inserted after every ReLU. For exact specifications of the parameters, the reader may check \url{https://github.com/chengtaipu/lowrankcnn}. We evaluated the performance of the models both with and without data augmentation. With data augmentation, the images are flipped horizontally with probability $0.5$ and translated in both directions by at most 1 pixel. Otherwise, we only subtract the mean of the images and normalize each channel.  The results are listed in Table \ref{tb:cifar_performance}.

The performance of the low-rank constrained versions of both networks are better than the baseline networks, with and without data augmentation. Notably, the low-rank NIN model outperforms the baseline NIN model by more than 1\%. And as far as we know, this is also better than previously published results.

We then study how the empirical performance and speedup change as we vary the rank $K$. We choose the CNN+Dropout as baseline model with data augmentation described above.  The results are listed in Table \ref{tb:cifar_speedup}.

\begin{table}
\small
\caption{Speedup and performance change. Performance change is relative to the baseline CNN+Dropout model with accuracy 87.71\%.}
\label{tb:cifar_speedup}
\begin{center}
\begin{tabular}{lllllllll}
\multicolumn{1}{c}{\bf LAYER}  & \multicolumn{1}{c}{\bf $K_1$}
&\multicolumn{1}{c}{\bf $K_2$} & \multicolumn{1}{c}{\bf $K_3$}
&\multicolumn{1}{p{1.8cm}}{\bf ACCURACY CHANGE} &\multicolumn{1}{p{1.8cm}}{\bf SPEEDUP (LAYER)} & \multicolumn{1}{p{1.6cm}}{\bf SPEEDUP (NET)}
&\multicolumn{1}{p{2.2 cm}}{\bf REDUCTIONS (WEIGHTS)}
\\ \hline\noalign{\smallskip}

 \multirow{3}{*}{First}& 4 & 64 & 256& +0.69\% & 1.20$\times$ & 2.91$\times$ &3.5$\times$\\
 &8 & 64 & 256& +0.85\% & 1.13$\times$ & 2.87$\times$ &1.8$\times$\\
 &12 & 64 & 256&  +0.94\% & 1.05$\times$ & 2.85$\times$ &1.2$\times$\\
\hline\noalign{\smallskip}
 \multirow{5}{*}{Second}
 	&12 & 8 & 256& -0.02\%  & 7.13$\times$ & 3.21$\times$ & 47.5$\times$ \\
    &12 & 16 & 256& +0.50\% & 6.76$\times$ & 3.21$\times$&23.8$\times$\\
 	&12 & 32 & 256& +0.89\% & 6.13$\times$ & 3.13$\times$&12.0$\times$\\
  	&12 & 64 & 256& +0.94\% & 3.72$\times$ & 2.86$\times$& 6.0$\times$\\
  	&12 & 128 & 256& +1.32\% & 2.38$\times$ & 2.58$\times$& 3.0$\times$\\
    &12 & 256 & 256& +1.40\% & 1.25$\times$ & 1.92$\times$& 1.5$\times$\\
\hline\noalign{\smallskip}
\multirow{5}{*}{Third}
& 12 & 64 & 8 & -2.25\% & 6.98 $\times$ & 3.11$\times$ & 52.5$\times$ \\
& 12 & 64 & 16 & +0.21\% & 6.89$\times$ & 3.11$\times$& 26.4$\times$\\
& 12 & 64 & 32 & +0.19\% & 5.82$\times$ & 3.10$\times$& 13.3$\times$\\
& 12 & 64 & 64 & +0.19\% & 3.74$\times$ & 2.96$\times$& 6.7$\times$\\
& 12 & 64 & 128 & +0.94\% & 2.38$\times$ & 2.86$\times$& 3.3$\times$\\
& 12 & 64 & 256 & +1.75\% & 1.31$\times$ & 2.30$\times$& 1.7$\times$\\
\end{tabular}
\end{center}
\end{table}

The number of parameters in the network can be reduced by a large factor, especially for the second and third layers. Up to $7\times$ speedup for a specific layer and  2-3$\times$ speedup for the whole network can be achieved. In practice, it is difficult for the speedup to match the theoretical gains based on the number of operations, which is roughly proportional to the reduction of parameters. The actual gain also depends on the software and hardware optimization strategies of convolutions. Our results in Table \ref{tb:cifar_speedup} are based on Nvidia Titan GPUs and Torch 7 with cudnn backend.

Interestingly, even with significant reductions in the number of parameters, the performance does not decrease much. Most of the networks listed in Table \ref{tb:cifar_speedup} even outperform the baseline model. Applying the low-rank constraints for all convolutional layers, the total number of parameters in the convolutional layers can be reduced by a large factor without degrading much performance. For example, with $K_1=12,K_2=16$ and $K_3=32$, the parameters in the convolutional kernels are reduced by 91\% and the relative performance is +0.25\%.

% The reason for this improvement is perhaps two fold. It could be that the low-rank structure serves as a regularizer as also noted by \cite{lebedev2014speeding,jaderberg2014speeding}. Indeed, we observed sometimes the training error of the low-rank CNNs are slightly bigger than the baseline CNN, but generalizes better. Indicating the low-rank constraint helps overcome the over-fitting issue. However, we observed that in some cases, the training errors of the low-rank CNNs are smaller than the baseline model too. Note that the expressive power of the baseline model is strictly bigger than the low-rank constrained counterparts\footnote{There is an extra bias term in the low-rank convolutional layer, but the effect of this bias term is small as indicated by their small magnitudes.}. This suggests low-rank constrained model gives a better starting point for the iterations, and reaches at a better local minimum. Such phenomenon is intriguing as it seems to suggests that the space of ``natural image patches'' have a low-rank structure.

Nevertheless, the parameters in the fully connected layers still occupy a large fraction. This limits the overall compression ability of the low-rank constraint. There are some very recent works focusing on reducing the parameters in the fully connected layers \citep{2015arXiv150906569N}, combining these techniques with the proposed scheme will be explored in future research.

\subsection{ILSVRC12}
ILSVRC12~\citep{ILSVRC15} is a well-known large-scale benchmark dataset for image classification. We adopt three famous CNN models, AlexNet~\citep{krizhevsky2012imagenet} (CaffeNet~\citep{jia2014caffe} as an variant), VGG-16~\citep{simonyan2014very}, and GoogLeNet~\citep{szegedy2014going} (BN-Inception~\citep{ioffe2015batch} as an variant) as our baselines. The CaffeNet and VGG-16 are directly downloaded from Caffe's model zoo and then fine-tuned on the training set until convergence, while the BN-Inception model is trained from scratch by ourselves.

The introduced low-rank decomposition is applied to each convolutional layer that has kernel size greater than $1\times 1$. Input images are first warped to $256\times 256$ and then cropped to $227\times 227$ or $224\times 224$ for different models. We use the single center crop during the testing stage, and evaluate the performance by the top-5 accuracy on the validation set. Detailed training parameters are available at \url{https://github.com/chengtaipu/lowrankcnn}.

\begin{figure}
\centering
\begin{minipage}[t]{0.48\textwidth}
\includegraphics[width=1.0\linewidth]{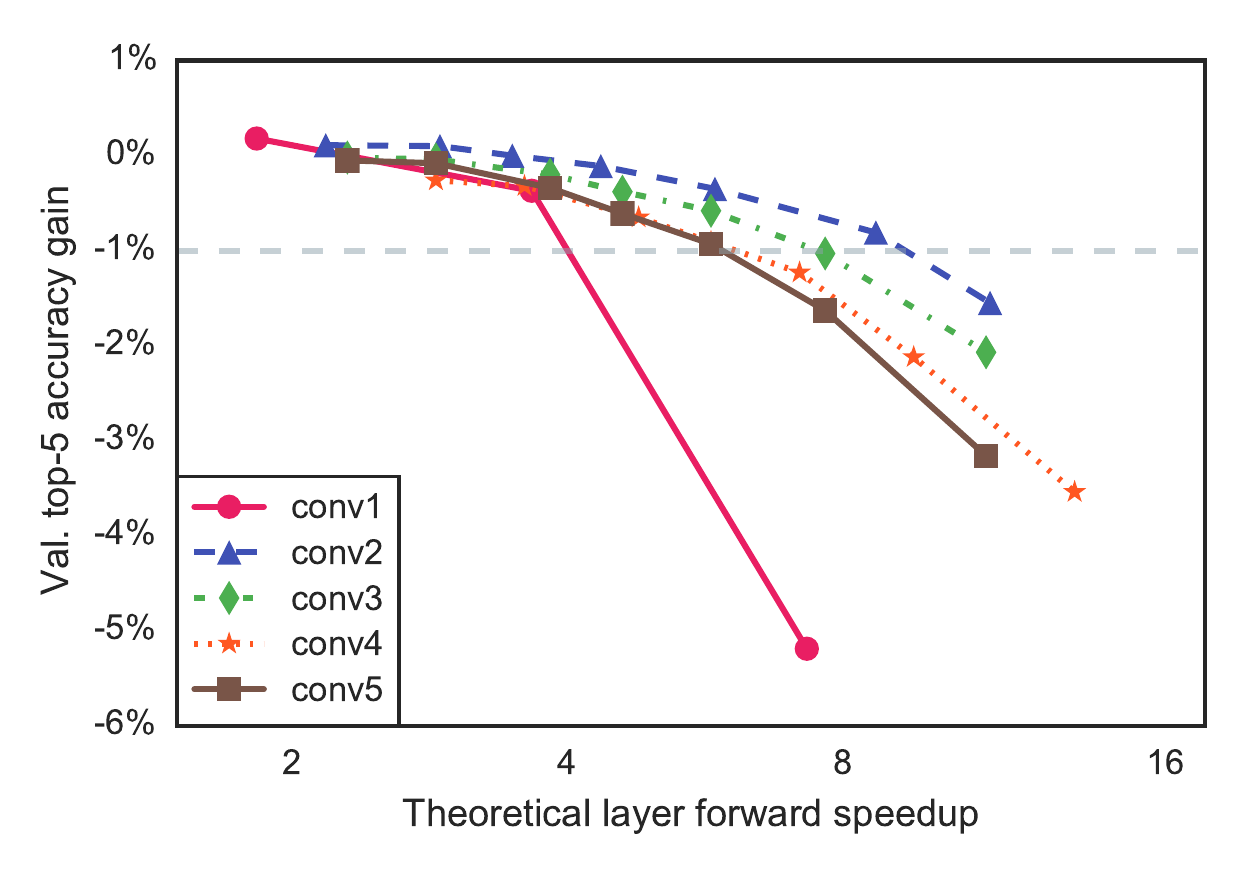}
\caption{The performance w.r.t. the theoretical layer speedup. Only the conv1-conv5 layers of the AlexNet are shown.}
\label{fig:cls_wrt_speedup}
\end{minipage}
\hfill
\begin{minipage}[t]{0.48\textwidth}
\includegraphics[width=1.0\linewidth]{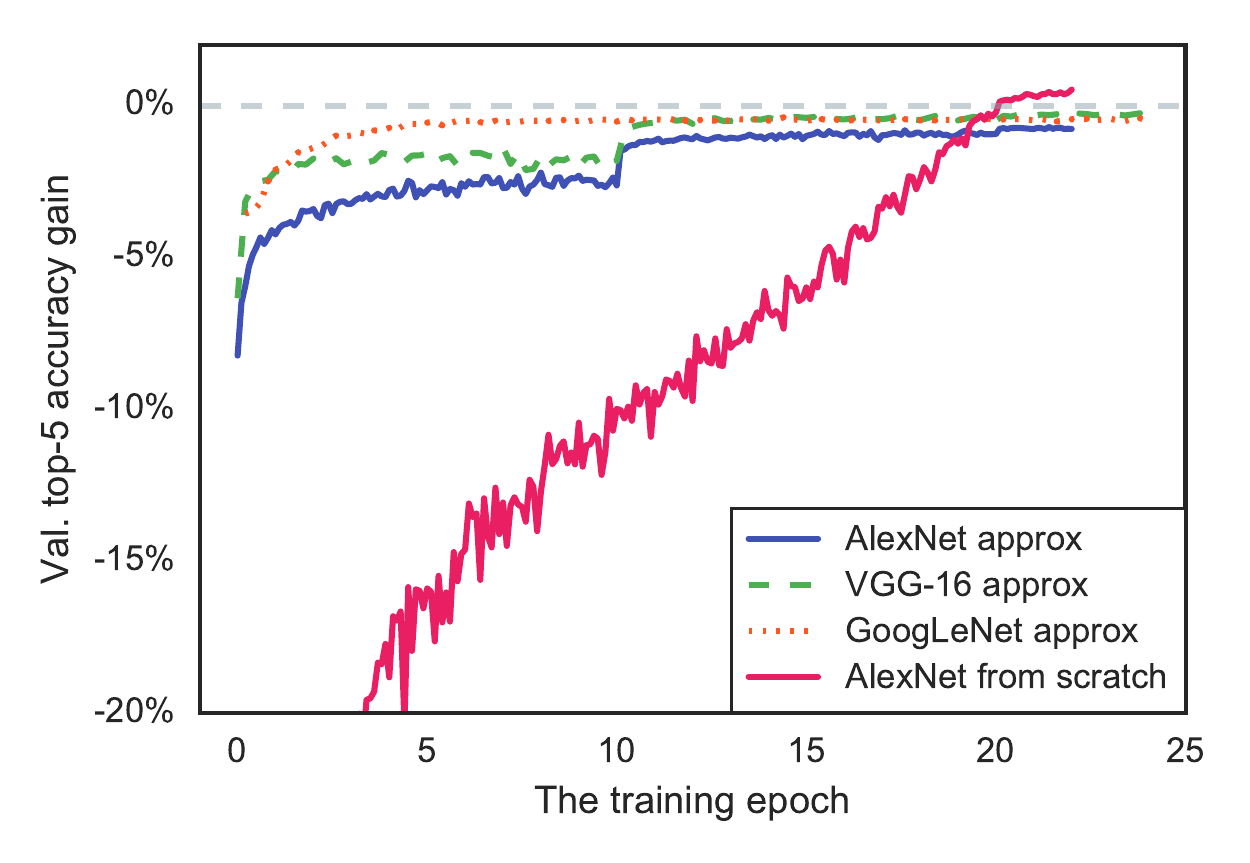}
\caption{The performance w.r.t. the fine-tuning epoch when using the proposed closed form solution as initialization.}
\label{fig:cls_wrt_epoch}
\end{minipage}
\end{figure}

As before, the hyper-parameter $K$ controls the trade-off between the speedup factor and the classification performance of the low-rank models. Therefore, we first study its effect for each layer, and then use the information to configure the whole low-rank model for better overall performance. We decompose a specific layer with a different $K$ each time, while keeping the parameters of all the other layers fixed. The performance after fine-tuning with respect to the theoretical layer speedup is demonstrated in Figure~\ref{fig:cls_wrt_speedup}. In general, we choose for each layer the value of $K$ that most accelerates the forward computation while does not hurt the performance significantly ($< 1\%$).  A more automatic way for choosing $K$ is based on Eigengap, such that the first $K$ eigenvectors account for 95\% of the variations. This is similar to choosing the number of principal components in PCA. The detailed low-rank model structures are listed in Table~\ref{tb:ilsvrc12_net}.

\begin{table}
\small
\centering
\caption{Low-rank models for ILSVRC12. For VGG-16, each convolution module contains two or three sub-convolutional layers. For GoogLeNet, each inception module contains one $3\times 3$ and two consecutive $3\times 3$ convolutional layers. Their corresponding $K$s are shown in a cell for brevity.}
\label{tb:ilsvrc12_net}
\subfloat[AlexNet]{
  \begin{tabular}{|c|c|}
  \hline
  Layer & $K$\\
  \hline
  conv1 & 8\\
  \hline
  conv2 & 40\\
  \hline
  conv3 & 60\\
  \hline
  conv4 & 100\\
  \hline
  conv5 & 200\\
  \hline
  \end{tabular}
  \hspace{.1em}
}
\subfloat[VGG-16]{
  \begin{tabular}{|c|c|}
  \hline
  Layer & $K$\\
  \hline
  conv1 & 5, 24\\
  \hline
  conv2 & 48, 48\\
  \hline
  conv3 & 64, 128, 160\\
  \hline
  conv4 & 192, 192, 256\\
  \hline
  conv5 & 320, 320, 320\\
  \hline
  \end{tabular}
  \hspace{.1em}
}
\subfloat[GoogLeNet]{
  \begin{tabular}{|c|c|c|c|}
  \hline
  Layer & $K$ & Layer & $K$\\
  \hline
  conv1 & 8 & inception(4b) & 64, 64, 80\\
  \hline
  conv2 & 48 & inception(4c) & 64, 64, 64\\
  \hline
  inception(3a) & 32, 32, 48 & inception(4d) & 64, 96, 96\\
  \hline
  inception(3b) & 32, 32, 48 & inception(4e) & 64, 128, 160\\
  \hline
  inception(3c) & 80, 32, 48 & inception(5a) & 128, 96, 128\\
  \hline
  inception(4a) & 32, 64, 80 & inception(5b) & 128, 96, 128\\
  \hline
  \end{tabular}
}
\end{table}

The proposed closed form solution provides the optimal data-independent initialization to the low-rank model. As indicated in Figure \ref{fig:cls_wrt_epoch}, there is a performance gap between the  low-rank models and their baselines at the beginning, but the performance is restored after fine-tuning. It is claimed in \citet{denton2014exploiting} that data-dependent criterion leads to better performance, we found that this is true upon approximation, but after fine-tuning, the difference between the two criteria is negligible ($<0.1\%$).

At last, we compare the low-rank models with their baselines from the perspective of classification performance, as well as the time and space consumption. The results are summarized in Table~\ref{tb:ilsvrc12_ret}. We can see that all the low-rank models achieve comparable performances. Those initialized with closed form weights approximation (cf. approximation rows in Table~\ref{tb:ilsvrc12_ret}) are slightly inferior to their baselines. While the low-rank AlexNet trained from scratch with BN could achieve even better performance. This observation again reveals that the low-rank CNN structure could have better discriminative power and generalization ability. On the other hand, both the running time and the number of parameters are consistently reduced. Note that the large gaps between the theoretical and the actual speedup are mainly due to the CNN implementations, and the current BN operations significantly slow down the forward computation. This suggests room for accelerating the low-rank models by designing specific numerical algorithms.
\begin{table}
\small
\centering
\caption{Comparisons between the low-rank models and their baselines. The theoretical speedup and weights reduction are computed concerning only the convolutional layers to be decomposed. While the actual speedup is based on the forward computation time of the whole net.}
\label{tb:ilsvrc12_ret}
\begin{tabular}{lcccc}
\multicolumn{1}{c}{\textbf{METHOD}} & \specialcell{\textbf{TOP-5 VAL.}\\\textbf{ACCURACY}} & \specialcell{\textbf{THEORETICAL}\\\textbf{SPEEDUP}} & \specialcell{\textbf{ACTUAL}\\\textbf{SPEEDUP}} & \specialcell{\textbf{WEIGHTS}\\\textbf{REDUCTION}}\\
\noalign{\smallskip}\hline\noalign{\smallskip}
AlexNet (original) & 80.03\% & $1\times$ & $1\times$ & $1\times$\\
Low-rank (cf. approximation) & 79.66\% & $5.27\times$ & $1.82\times$ & $5.00\times$\\
Low-rank (from scratch with BN) & 80.56\% & $5.24\times$ & $1.09\times$ & $4.94\times$\\
\hline\noalign{\smallskip}
VGG-16 (original) & 90.60\% & $1\times$ & $1\times$ & $1\times$\\
Low-rank (cf. approximation) & 90.31\% & $3.10\times$ & $2.05\times$ & $2.75\times$\\
\hline\noalign{\smallskip}
GoogLeNet (original) & 92.21\% & $1\times$ & $1\times$ & $1\times$\\
Low-rank  (cf. approximation) & 91.79\% & $2.89\times$ & $1.20\times$ & $2.84\times$\\
\end{tabular}
\end{table}

\section{Discussion}
In this paper, we explored using tensor decomposition techniques to speedup convolutional neural networks. We have introduced a new algorithm for computing the low-rank tensor decomposition and a new method for training low-rank constrained CNNs from scratch. The proposed method is evaluated on a variety of modern CNNs, including AlexNet, NIN, VGG, GoogleNet with success. This gives a strong evidence that low-rank tensor decomposition can be a generic tool for speeding up large CNNs.

On the the other hand, the interesting fact that the low-rank constrained CNNs sometimes outperform their non-constrained counterparts points to two things. One is the local minima issue. Although the expressive power of low-rank constrained CNNs is strictly smaller than that of the non-constrained one, we have observed in some cases that the former have smaller training error. This seems to suggest the low-rank form helps the CNNs begin with a better initialization and settles at a better local minimum. The other issue is over-fitting. This is shown by the observation that in many cases the constrained model has higher training error but generalizes better. Overall, this suggests room for improvement in both the numerical algorithms and the regularizations of the CNN models.
\section*{Acknowledgments}
This work is supported in part by the 973 project 2015CB856000 of the
Chinese Ministry of Science and Technology and the DOE grant DE-SC0009248.

\bibliography{iclr2016_conference}
\bibliographystyle{iclr2016_conference}
\subsection*{Appendix}
\subsection*{Proof of Theorem 1}

\begin{proof}
Consider the following minimization problem:
\begin{equation}
\begin{aligned}
 (P2)\quad\quad E_2(\tilde{W}) &:= \|\tilde{W}-W\|_F^2 \\
\text{subject to} \quad &\text{Rank}(\tilde{W}) \leq K.
\end{aligned}
\label{eq:p2}
\end{equation}
Let $(\H^*,\V^*)$ be a solution to (P1), then we can construct a solution to $(P2)$ as follows:
\[
  \tilde{W} = \sum_{k=1}^K \begin{bmatrix} \V_k^1\\ \V_k^2 \\ \vdots \\\V_k^C\end{bmatrix}
  \begin{bmatrix} \H^k_1, \H^k_2, \cdots, \H^k_N\end{bmatrix}.
\]
Because of the separability of the Frobenius norm,
\[
  E_1(\H^*,\V^*)=E_2(\tilde{W}).
\]
Moreover, as $\text{Rank}(\tilde{W})\leq K$, hence $\tilde{W}$ is feasible for $(P2)$. We have
\begin{equation}
 	E_2 (W^*)\leq E_1(\H^*,\V^*)=E_2(\tilde{W}) ,
\label{eq:e2}
\end{equation}
where $W^*$ is any solution to $(P2)$.

On the other hand, let $W^*$ be a solution to $(P2)$, then we construct a solution $(\hat{\H},\hat{\V})$ to $(P1)$ as \eqref{eq:soln1}. Hence
\[
     E_1(\H^*,\V^*)\leq E_1(\hat{\H}, \hat{\V}) .
\]
Together with \eqref{eq:e2},
\begin{equation}
       E_1(\hat{\H},\hat{\V}) = E_2(\W^*) = E_1(\H^*,\V^*).
\end{equation}
 We have proved $(\hat{\H}, \hat{\V})$ is a solution to $(P1)$.
\end{proof}
\subsection*{Hardness of the data-dependent approximation}
Using the data-dependent criterion, the minimization problem is:
\begin{equation}
  E(\H,\V) := \sum_{i=1}^{M} \sum_{n=1}^N \sum_{c=1}^C \|\W_n^c*\Z_i^c - \sum_{k=1}^K \H_n^k(\V_k^c)^T * \Z_i^c \|_F^2.
\end{equation}
For fixed stride $s$, define the linear map $P_m: \R^{X\times Y} \mapsto \R^{d\times d}$, $P_m(z)$ samples the $m$-th $d\times d$ patch from $z\in \R^{X\times Y}$ followed by flipping the patch horizontally and vertically. Then
\[
  \sum_{c} \|\W_n^c *\Z_n^c \|_F^2 = \sum_{m,c} \langle \W_n^c, P_m\Z_{i}^c \rangle^2.
\]

Let
\[
	Z_{im} = \begin{bmatrix}  P_m \Z_i^1 \\ P_m \Z_i^2 \\ \vdots \\ P_m\Z_i^c  \end{bmatrix}\otimes \underbrace{(1,1,\cdots, 1)}_{N}.
\]
Similar as in Criterion 1, the approximation problem is equivalent to the following minimization program:
\begin{equation}
\begin{aligned}
    E(\tilde{W})&:= \sum_{i,m} \|(W-\tilde{W})\circ Z_{im}\|_F^2 \\
    \text{subject to} \quad & \quad \text{Rank}(\tilde{W}) \leq K,
\end{aligned}
\end{equation}
where $\circ$ is the Hadamard product.

This is a weighted low-rank approximation problem:
\begin{equation}
\begin{aligned}
    E(\tilde{W})&:= \sum_{ij} G_{ij}(W_{ij} -\tilde{W}_{ij})^2\\
    \text{subject to} \quad & \quad \text{Rank}(\tilde{W}) \leq K,
\end{aligned}
\end{equation}
where $G = \sum_{im}Z_{im}\circ Z_{im}$.

Although it appears very similar to the problem in Criterion 1, which has a closed form solution, this is much more difficult to solve except for a few special cases. (E.g., when the weight matrix is identity or has rank one.) In fact, it can be proved that this problem is NP-hard \citep{gillis2011low}.

\end{document}